\PassOptionsToPackage{sort&compress}{natbib}
\documentclass[twocolumn]{templates/arxiv/fairmeta}

\usepackage{amsmath}
\usepackage{amssymb}
\usepackage{mathtools}
\usepackage{bm}
\usepackage{algorithm}
\usepackage{algpseudocode}
\usepackage[most]{tcolorbox}
\usepackage[dvipsnames,table,xcdraw]{xcolor}
\usepackage{booktabs}
\usepackage{array}
\usepackage{tabularx}
\usepackage{ragged2e}
\usepackage{makecell}
\usepackage{multirow}
\usepackage{tikz}
\usepackage{pgfplots}
\usepackage{wrapfig}
\usetikzlibrary{pgfplots.groupplots, matrix}
\pgfplotsset{compat=1.18}
\usepackage{enumitem}
\usepackage{comment}
\usepackage{graphicx}
\usepackage{colortbl}
\usepackage{anyfontsize}
\usepackage{xspace}

\definecolor{navyblue}{RGB}{30,130,255}
\definecolor{citecolor}{RGB}{30,130,255}
\definecolor{lightgray}{gray}{0.9}

\newcommand{\method}{IDWorld\xspace}

\providecommand{\eg}{\textit{e.g.}}
\providecommand{\ie}{\textit{i.e.}}

\newcolumntype{L}[1]{>{\RaggedRight\arraybackslash}p{#1}} %
\newcolumntype{Y}{>{\RaggedRight\arraybackslash}X}        %

\makeatletter
\newskip\IDWparfill
\patchcmd{\@trivlist}{\parfillskip \@flushglue}{\parfillskip \IDWparfill}{}{}
\makeatother
\setcitestyle{numbers,square,comma}

\usepackage{adjustbox}
\newcommand{\colfigwidth}{\linewidth}
\newcommand{\fittable}[1]{\resizebox{\linewidth}{!}{#1}}
\newcommand{\widetable}[1]{\adjustbox{max width=\linewidth}{#1}}
\newcommand{\coltable}[1]{\adjustbox{max width=\linewidth}{#1}}

\titleformat{\paragraph}[runin]{\normalfont\normalsize\sffamily\bfseries}{}{0pt}{}
\titlespacing*{\paragraph}{0pt}{\parskip}{0.5em}
\DeclareTextFontCommand{\textbfrm}{\bfseries\rmfamily}
\AtBeginEnvironment{tabular}{\let\textbf\textbfrm}

\title{Identity World Model: Recognizing People in an Imagined Latent World}

\author[1]{Dingqiang Ye}
\author[1]{Kartik Narayan}
\author[1]{Vishal M. Patel}

\affiliation[1]{Johns Hopkins University}

\abstract{Whole-body human recognition has advanced rapidly with large vision models.
However, existing methods remain fragile under real-world variations, as they tie identity to transient cues such as clothing, texture, pose, or viewpoint.
We argue that a robust recognizer should not merely learn to ignore these cues, but imagine how a person would be represented if they changed.
To this end, we propose \textbf{\method}, an identity world model: just as a latent world model predicts the features of a scene after an action, \method predicts how a person would be represented under a changed condition.
In a single forward pass, \method imagines these counterfactual features under shared canonical conditions, such as the same outfit, a fixed view, or a bare silhouette, without rendering any pixels.
Compared with explicit imagination in pixels, this latent imagination keeps nearly the same accuracy while being over 11$\times$ faster, at 5.9\,ms per frame.
Since each imagined condition has its own strengths, \method then fuses them adaptively, weighting each condition for every probe-gallery pair.
Extensive experiments show that \method achieves state-of-the-art results on CCPG and CCGR-MINI, outperforms the compared methods on MEVID, and generalizes best to CASIA-B* and SUSTech1K, with the best mAP on CCVID.
These results suggest identity world models offer a simple yet effective path to robust recognition.
Code will be released.
}

\correspondence{Dingqiang Ye}

\begin{document}

\maketitle
\begin{figure*}[t]
\centering
\includegraphics[width=0.95\linewidth]{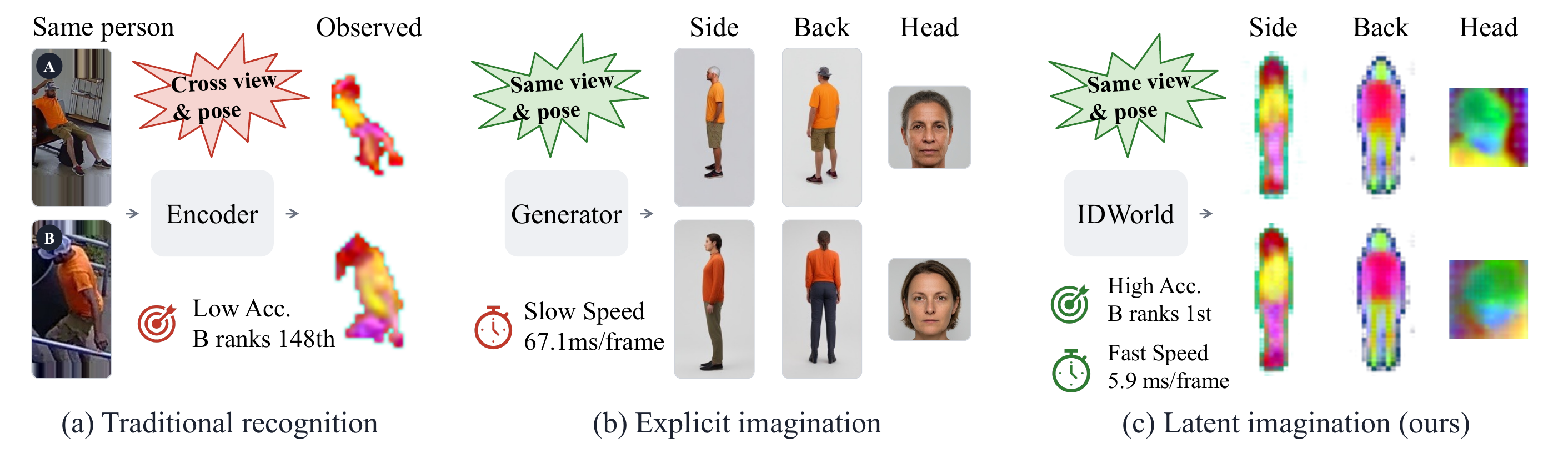}
\caption{Comparison of recognition paradigms on one person from MEVID~\cite{davila2023mevid}, who sits in A and walks in B under different views. (a)~Traditional recognition naively matches observed features, ranking B only 148th. (b)~Explicit imagination renders shared conditions in pixels, spending an expensive 67.1\,ms per frame. (c)~Our \method efficiently imagines them in latent space, ranking B first in only 5.9\,ms.}
\label{fig:teaser}
\end{figure*}

\raggedbottom
\clubpenalty=10000
\widowpenalty=10000
\section{Introduction}
\label{sec:intro}

Whole-body human recognition identifies a person from the appearance and movement of the entire body; it needs neither a frontal face nor the subject's cooperation and works at a distance and across cameras~\cite{sepas2023gaitsurvey,ye2022reidsurvey}, making it vital for public safety and long-term tracking.
It spans gait recognition~\cite{fan2023opengait,ye2024biggait,ye2025biggergait}, which identifies people by their walk, and person re-identification (ReID)~\cite{chen2023solider,li2023clipreid,kim2025sapiensid}, which matches people by appearance; both have advanced rapidly with large vision models and human-centric pretraining.

Yet recognition in the wild remains difficult, as observations of the same person vary greatly.
New clothing changes colors and textures, and a long coat alters the body outline and hides the legs~\cite{li2023ccpg,qian2020ltcc}; viewpoint decides which body parts frontal, lateral, and overhead cameras see~\cite{yu2006casiab,zou2024ccgr}; and pose reshapes the body as a person walks, turns, or carries objects~\cite{davila2023mevid}.
As these factors compound, two observations of the same person may share little.

Existing methods cope with this variability in three ways, each with its own weakness.
The first removes appearance altogether, relying on silhouettes and skeletons in gait recognition~\cite{chao2019gaitset,teepe2021gaitgraph,fu2023gpgait,fan2024skeletongait} or on contours, body parsing, and 3D shape in cloth-changing ReID~\cite{yang2021prcc,hong2021fsam,chen2021shape3d}; such inputs discard much discriminative detail, yet remain sensitive to coat outlines and viewpoint.
The second learns invariance from data through adversarial objectives, disentangled representations, or causal intervention~\cite{yu2017gaitgan,gu2022cal,yang2023aim}, but this invariance covers only the combinations seen in training.
The third builds on large pretrained models~\cite{li2023clipreid,ye2024biggait,ye2025biggergait,kim2025sapiensid}, which are highly discriminative but handle these changes only implicitly, so their robustness is hard to control or interpret.
In short, all three learn what to ignore: variation is noise, not change to anticipate (\cref{fig:teaser}a).

Humans, by contrast, anticipate such changes: seeing a familiar person in a new coat or from behind, we readily picture how they would look in their usual clothes or from the front, and compare them under the same conditions.
World models bring this ability to machines by predicting the outcome of an action~\cite{ha2018worldmodels,hafner2020dreamer,lecun2022path,bruce2024genie}; in recognition, each shared condition, \eg, the same outfit or view, is an action: the model predicts how a person would be represented under it, so two observations are compared on equal terms.
Modern image editors can already re-dress and re-pose a person~\cite{brooks2023instructpix2pix,labs2025fluxkontext,wu2025qwenimage}, and earlier work rendered faces and bodies in canonical poses for a similar purpose~\cite{tran2017drgan,huang2017tpgan,qian2018pngan}, suggesting that this vision is within reach.

Turning this idea into a practical imagination-based recognizer, however, raises two challenges.
(1)~\textbf{High latency}: explicit imagination in pixels costs tens of milliseconds per frame even with a 4-step distilled editor, which is prohibitive for videos and large galleries (\cref{fig:teaser}b).
(2)~\textbf{Varying utility}: each condition has its own strengths, \eg, the silhouette and the same outfit are robust to clothing changes, while the side and back views bridge camera and posture gaps, so the right condition depends on each probe-gallery pair.
Such a recognizer should thus be fast and selective.

To address these challenges, we propose \method, an identity world model that predicts how a person would be represented under changed conditions.
\method has three stages: explicit imagination provides supervision, latent imagination learns from it, and adaptive fusion weights the imagined identity features.
First, an image editor~\cite{wu2025qwenimage} renders each training person under five canonical conditions, and a condition expert is trained for each.
The conditions complement each other: the silhouette and the same outfit address clothing changes, the side and back views address viewpoint and pose changes, and the head close-up adds facial cues (\cref{fig:teaser}c).
Second, to cut latency, an imagination head takes the condition as its action and, like latent world models~\cite{assran2023ijepa,zhou2025dinowm,assran2025vjepa2}, regresses the experts' early feature maps from RGB alone, over 11$\times$ faster than explicit imagination.
Third, to handle varying utility, adaptive fusion weights each condition for every probe-gallery pair according to how the two observations differ.
Overall, \method is a restricted world model~\cite{assran2023ijepa} that models a person's appearance uncertainty from a partially observed condition.

On six gait and video-based ReID benchmarks~\cite{li2023ccpg,zou2024ccgr,yu2006casiab,shen2023lidargait,gu2022cal,davila2023mevid}, \method outperforms prior methods in nearly all within-domain and cross-domain settings.
Our analysis further reveals three insights: (1)~latent imagination is cheap while keeping most of the gain of explicit imagination, (2)~early feature maps are better imagination targets than final embeddings, and (3)~direct regression beats flow matching from noise as the training objective.

In summary, the main contributions are as follows:
\begin{itemize}[leftmargin=*,topsep=2pt,itemsep=1pt]
  \item We present the first identity world model, a pioneering and practical paradigm for whole-body human recognition, which anticipates rather than suppresses variation by imagining people under shared conditions in latent space.
  \item We instantiate this paradigm as \method, which runs over 11$\times$ faster than explicit imagination with nearly the same accuracy and sets new state-of-the-art results on six gait and video-based ReID datasets in most settings.
  \item We provide in-depth analyses of what to imagine, how to learn it, and how to fuse it, offering insights that can inspire future research on imagination-based recognition.
\end{itemize}

\section{Related work}
\label{sec:related}

\begin{figure*}[t]
  \centering
  \includegraphics[width=0.94\linewidth]{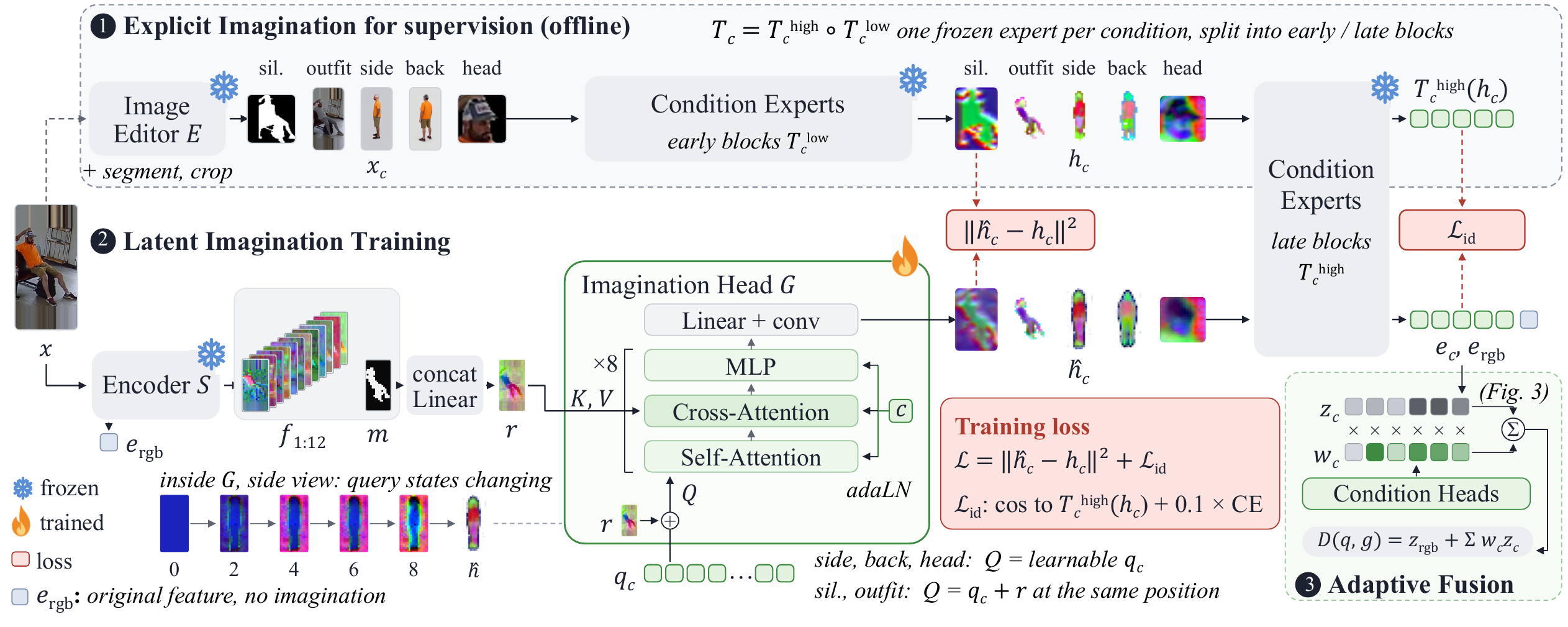}
  \caption{Overview of \method. (1)~Offline, each training input $x$ is rendered under every canonical condition, and the early blocks $T_c^{\mathrm{low}}$ of frozen experts extract the targets $h_c$. (2)~From the frozen encoder $S$, the imagination head $G$ predicts all $\hat{h}_c$ in a single forward pass. (3)~Adaptive fusion weights the imagined identity features for each probe-gallery pair (details in \cref{fig:fusion}). At test time, only RGB is used.}
  \label{fig:overview}
\end{figure*}

\paragraph{Whole-body human recognition.}
Whole-body human recognition advances along two lines, gait recognition and person ReID, both striving for robustness to clothing, viewpoint, and pose.
Gait recognition progressed from model-based methods on body pose~\cite{liao2020posegait,teepe2021gaitgraph,fu2023gpgait,fan2024skeletongait,guo2025gaitcontour} and appearance-based methods on silhouettes~\cite{wu2017crossview,chao2019gaitset,fan2020gaitpart,lin2021gaitgl,huang2021gait3dlocal,fan2023opengait,fan2023deepgaitv2,guo2024gader} to disentangled and dense 3D representations~\cite{zhang2019gaitnet,zhang2022gaitnet,zheng2022gait3d,wang2025vmgait,wang2026meshgait} and RGB-based methods built on large vision models~\cite{ye2024biggait,ye2025biggergait}, on benchmarks of growing scale and difficulty~\cite{yu2006casiab,takemura2018oumvlp,zhu2021grew,zheng2022gait3d,shen2023lidargait,li2023ccpg,zou2024ccgr}.
Person ReID evolved from part-based, multi-granularity, and transformer models~\cite{sun2018pcb,wang2018mgn,zhou2019osnet,luo2019bot,he2021transreid} to human-centric pretraining~\cite{fu2021luperson,zhu2022pass,chen2023solider,tang2023humanbench,ci2023unihcp,khirodkar2024sapiens,kim2025sapiensid} and vision-language models~\cite{li2023clipreid}, and recent analyses probe which attributes their embeddings encode~\cite{pal2026attribe}.
Cloth-changing ReID exploits clothing-irrelevant cues such as contours, body shape, and 3D structure~\cite{yang2021prcc,hong2021fsam,chen2021shape3d,liu2023invarreid}, or suppresses clothing bias with adversarial and causal objectives~\cite{gu2022cal,yang2023aim}, on benchmarks that vary clothing, viewpoint, altitude, and range~\cite{zheng2015market,wei2018msmt17,huang2019celebreid,qian2020ltcc,wan2020vcclothes,gu2022cal,davila2023mevid,cornett2023briar}.
Whole-body systems combine face, body, and gait~\cite{guo2022multimodal,wang2024hypergait,wang2025unigait,wang2026combogait,liu2024farsight}; classic score fusion normalizes and sums scores with fixed rules~\cite{jain2005score,cormack2009rrf}, and FusionAgent~\cite{zhu2026fusionagent} lets an agent choose the recognizers to combine for each sample.
Yet all of them suppress variation in the observation at hand, whereas \method anticipates it by imagining people under shared conditions and fusing them adaptively per probe-gallery pair.

\paragraph{Imagination for recognition.}
A long line of work canonicalizes observations by imagining a person under another condition.
In pixel space, faces are rotated to a frontal view~\cite{hassner2015frontalization,tran2017drgan,huang2017tpgan,yin2017ffgan}, pedestrians are re-rendered in new or canonical poses~\cite{ma2017pg2,liu2018posetransfer,qian2018pngan,ge2018fdgan}, and gait representations are transformed across views~\cite{makihara2006vtm,yu2017gaitgan}; generative models also synthesize training images with new appearances and structures~\cite{zheng2017unlabeled,zheng2019dgnet}, and modern diffusion models and editors can re-dress, re-pose, or animate a person from a single image~\cite{rombach2022ldm,zhang2023controlnet,brooks2023instructpix2pix,hu2024animateanyone,labs2025fluxkontext,wu2025qwenimage}.
Such explicit imagination, however, is too slow for large galleries.
In feature space, DREAM~\cite{cao2018dream} and later work~\cite{stanishev2025frontalization} map profile-face features to frontal ones, and feature augmentation transfers intra-class variations~\cite{yin2019featuretransfer,wang2019isda} or simulates clothing changes~\cite{han2023ccfa} during training; these methods are efficient, but each targets a single factor with a dedicated mapping, mostly at training time.
\method unifies both lines: renders supervise an imagination head that predicts all conditions in a single forward pass, keeping each separate for adaptive fusion.

\paragraph{World models and latent imagination.}
World models, from compact models of simple environments~\cite{ha2018worldmodels,hafner2020dreamer,hafner2023dreamerv3} to generative simulators of interactive and physical worlds~\cite{yang2024unisim,bruce2024genie,agarwal2025cosmos}, predict how the world would evolve so that agents can plan before acting.
Predicting in a learned latent space rather than in pixels has proven both efficient and effective~\cite{hafner2020dreamer,lecun2022path,assran2023ijepa,zhou2025dinowm,assran2025vjepa2}.
In robotics, world action models jointly imagine future observations and predict actions~\cite{cheang2024gr2,li2025uva,yuan2026fastwam}, and Fast-WAM~\cite{yuan2026fastwam} shows that their benefit largely persists even without imagining future frames in pixels at test time.
A related line hallucinates information unavailable at test time, as in learning with privileged information, knowledge distillation, modality hallucination, and cross-modal distillation~\cite{vapnik2009lupi,hinton2015distilling,romero2015fitnets,hoffman2016hallucination,gupta2016crossmodal}.
\method extends world models to identity: rather than an environment under actions or a missing modality, it imagines a person under canonical conditions in latent space.

\section{Method}
\label{sec:method}

In this section, we first outline \method (\cref{sec:overview}, \cref{fig:overview}) and then present its three stages: explicit imagination (\cref{sec:explicit}), latent imagination (\cref{sec:latent}), and adaptive fusion (\cref{sec:fusion}).
The first stage supervises the second, and inference runs the last two in a single forward pass (\cref{sec:inference}).

\subsection{Overview}
\label{sec:overview}
\method is an identity world model: from a single observation of a person, it predicts how the person would be recognized under every canonical condition.

\noindent\textbf{Input.}
An RGB image or video tracklet $x$ of a person.

\noindent\textbf{Output.}
Identity features and the observed condition of $x$.
The features are its original $e_{\mathrm{rgb}}$ and an imagined $e_c$ for each canonical condition $c \in \mathcal{C}$, \ie, a bare silhouette, the same outfit (plain gray), side and back views, and a head close-up, describing the person as if observed under $c$, the action of a world model.
The observed condition $o$ is how $x$ was captured: its camera elevation, walking direction, and outfit.
The features identify the person, and the observed condition guides their adaptive fusion.

\noindent\textbf{Overall mapping.}
Formally, \method implements
\begin{equation}
  x \mapsto \big(e_{\mathrm{rgb}},\, \{e_c\}_{c \in \mathcal{C}},\, o\big).
  \label{eq:world}
\end{equation}

\subsection{Explicit imagination for supervision}
\label{sec:explicit}
Explicit imagination provides the supervision for latent imagination in three steps (\cref{fig:overview}, stage 1).

\noindent\textbf{Rendering.}
The five conditions are rendered in two ways.
For the same outfit, side view, and back view, an image editor $E$~\cite{wu2025qwenimage} re-dresses the person in plain gray or turns them around with a fixed prompt shared by all identities and datasets; for the silhouette and the head close-up, $E$ stands for segmentation and keypoint-based cropping.
Each observation $x$ thus yields a render under every condition $c$,
\begin{equation}
  x_c = E(x, c).
  \label{eq:render}
\end{equation}

\noindent\textbf{Condition experts.}
To recognize people in these renders, we train a condition expert $T_c$ for each condition $c$: a silhouette recognizer~\cite{ye2026silhouette} for the silhouette and an RGB recognizer~\cite{ye2025biggergait} for the other four conditions.

\noindent\textbf{Imagination targets.}
We then split each frozen expert into early and late blocks, \mbox{$T_c = T_c^{\mathrm{high}} \circ T_c^{\mathrm{low}}$}, and take its early feature maps as the imagination target,
\begin{equation}
  h_c = T_c^{\mathrm{low}}(x_c).
  \label{eq:target}
\end{equation}
\cref{tab:ablation}a details the choice of split.
All targets are computed once, offline, and are never needed at test time.

\subsection{Latent imagination}
\label{sec:latent}
Latent imagination learns to reproduce these targets from RGB alone in a single forward pass (\cref{fig:overview}, stage 2).

\noindent\textbf{Frozen encoder.}
A frozen encoder $S$ reads the input and provides the context for imagination.
It is a DINOv2-S~\cite{oquab2024dinov2} backbone, on whose layers an RGB recognizer~\cite{ye2025biggergait} gives the original feature $e_{\mathrm{rgb}}$.
As these layers carry complementary cues~\cite{ye2025biggergait}, we concatenate the patch tokens of all 12 layers, $f_{1:12}$, with a self-supervised human mask $m$~\cite{ye2024biggait} and project them into the context $r = \mathrm{Linear}([f_{1:12};\, m])$.

\noindent\textbf{Imagination head.}
An imagination head $G$ predicts the target of each condition from this context,
\begin{equation}
  \hat{h}_c = G(Q_c, r, c),
  \label{eq:imagine}
\end{equation}
where the queries $Q_c$ align one-to-one with the positions of $h_c$.
For the side view, back view, and head, whose layout differs from the input, $Q_c$ consists of learnable queries $q_c$; for the silhouette and same outfit, which keep the input pose, the token of $r$ at the same position is added to $q_c$.
$G$ is a DiT-style transformer~\cite{peebles2023dit} of 8 blocks, each applying self-attention, cross-attention to $r$, and an MLP, modulated by adaLN on an embedding of $c$; a linear layer with a residual convolution outputs $\hat{h}_c$ in the shape of $h_c$.

\noindent\textbf{Training objective.}
The late expert blocks read the imagined feature maps out into the imagined identity feature $e_c = T_c^{\mathrm{high}}(\hat{h}_c)$, and $G$ is trained with a regression loss and an identity loss for every condition,
\begin{align}
  \mathcal{L} &= \sum\nolimits_{c \in \mathcal{C}} \big( \| \hat{h}_c - h_c \|_2^2 + \mathcal{L}_{\mathrm{id}}^{c} \big), \label{eq:loss} \\
  \mathcal{L}_{\mathrm{id}}^{c} &= 1 - \cos\big(e_c,\, T_c^{\mathrm{high}}(h_c)\big) + 0.1\, \mathrm{CE}(e_c, y), \label{eq:idloss}
\end{align}
where features are normalized per condition, $y$ is the identity label, and $\mathrm{CE}$ is the cross-entropy of the expert's classifier.
The identity loss keeps the imagined identity feature recognizable by its expert, and direct regression suits targets that are nearly determined by the input; \cref{tab:ablation}b compares it with flow matching from Gaussian noise~\cite{lipman2023flowmatching}.

\subsection{Adaptive fusion}
\label{sec:fusion}

\begin{figure}[t]
  \centering
  \includegraphics[width=\colfigwidth]{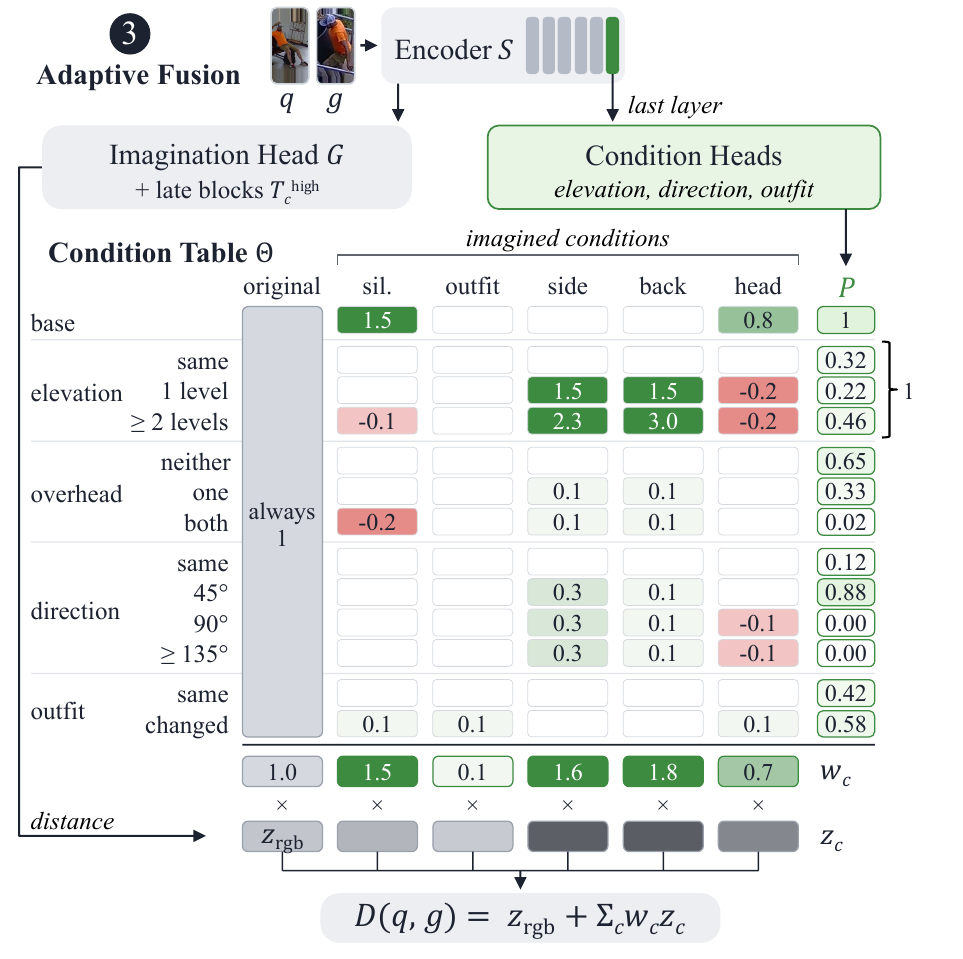}
  \caption{Adaptive fusion. The condition table $\Theta$ weights each imagined condition by how the probe $q$ and gallery $g$ differ. For this pair, seen from different elevations and directions, it raises the side and back views from 0 to 1.6 and 1.8, helping recognize the person across this large viewpoint gap.}
  \label{fig:fusion}
\end{figure}

Each imagined condition has its own strengths, so adaptive fusion weights the conditions for every pair of a probe $q$ and a gallery $g$ by how the two differ (\cref{fig:fusion}).

\noindent\textbf{Condition heads.}
Three linear heads on the last-layer features of the frozen DINOv2-S predict the observed condition $o$ of each observation: the camera elevation in five levels from eye level to overhead, the walking direction in five classes from front to back, and an outfit embedding.
For a pair, they give the probability of each state of four relations, the rows of \cref{fig:fusion}: the elevation gap, overhead views, the direction gap, and the outfit change.
Each gap state sums the probabilities of all class pairs that yield it, and the outfit change uses the cosine similarity of the two embeddings.
They are pre-trained with labels: views for the elevation and direction~\cite{zou2024ccgr}, and outfit pairs for the outfit~\cite{li2023ccpg,davila2023mevid}.

\noindent\textbf{Condition table.}
The condition table $\Theta$ turns these probabilities into a weight $w_c$ for each imagined condition.
Its columns are the conditions, and its rows are a base row and the states of the four relations.
The base row gives each condition a default weight, and every other row raises or lowers it by its entry times the probability $P$ of that state; the sum of each column, clipped at zero, is $w_c$.
The entries follow common sense: they are symmetric in $q$ and $g$ and monotonic in each gap, viewpoint gaps raise the side and back views, and outfit changes raise the clothing-independent conditions.
They are fit on training identities by coordinate ascent to maximize retrieval accuracy, starting from a predefined table and pulled toward it by an L1 penalty.

\noindent\textbf{Fused distance.}
The fused distance adds the weighted distances of all conditions to that of the original feature,
\begin{equation}
  D(q,g) = z_{\mathrm{rgb}} + \sum\nolimits_{c \in \mathcal{C}} w_c\, z_c,
  \label{eq:fusion}
\end{equation}
where $z_c$ is the Euclidean distance between the imagined identity features $e_c$ of $q$ and $g$, z-scored~\cite{jain2005score} over the gallery of $q$ to share one scale, and $z_{\mathrm{rgb}}$ is its counterpart on $e_{\mathrm{rgb}}$.

\subsection{Inference}
\label{sec:inference}
At test time, \method recognizes people from RGB alone (\cref{fig:overview}).
Each observation $x$ needs a single forward pass.
The frozen encoder $S$ gives the original feature $e_{\mathrm{rgb}}$.
The imagination head $G$ predicts the early feature maps of all conditions, and the late expert blocks read them out into $e_c$.
The condition heads predict the observed condition $o$.
These outputs form the mapping in \cref{eq:world}.
For each probe-gallery pair, the condition table weights the conditions by the observed conditions of $q$ and $g$.
\cref{eq:fusion} then gives the distance $D(q,g)$ for ranking.
No image is rendered, so inference is fast enough for videos and large galleries.

\section{Experiments}
\label{sec:experiments}
We evaluate \method on six gait and video-based ReID benchmarks, both within and across domains.
We compare \method with state-of-the-art methods (\cref{sec:exp_sota}) and then analyze its design quantitatively and qualitatively (\cref{sec:exp_ablation}): latent versus explicit imagination, the imagination target and training objective, the role of each condition, adaptive fusion, and the visualization of imagined feature maps.

\begin{table*}[t]
  \centering
  \caption{Within-domain gait recognition (\%): Rank-1 on the four CCPG settings and the mean, and Rank-1 (R1) and mAP on \mbox{CCGR-MINI}. CL, UP, DN, BG: outfit, top, bottom, and bag changes. Sil.: silhouette; Ske.: skeleton; Par.: parsing. LVM: large vision model.}
  \label{tab:sota_gait}
  \small
  \setlength{\tabcolsep}{4.6pt}
  \renewcommand{\arraystretch}{0.92}
  \widetable{%
  \begin{tabular}{llcc ccccc cc}
    \toprule
    \multirow{2}{*}{Input} & \multirow{2}{*}{Method} & \multirow{2}{*}{LVM} & \multirow{2}{*}{Venue} & \multicolumn{5}{c}{CCPG~\cite{li2023ccpg}} & \multicolumn{2}{c}{CCGR-MINI~\cite{zou2024ccgr}} \\
    \cmidrule(lr){5-9} \cmidrule(lr){10-11}
    & & & & CL & UP & DN & BG & Mean & R1 & mAP \\
    \midrule
    \multirow{4}{*}{Sil.} & GaitBase~\cite{fan2023opengait} & - & CVPR'23 & 71.6 & 75.0 & 76.8 & 78.6 & 75.5 & 27.0 & 24.9 \\
     & DeepGaitV2~\cite{fan2023deepgaitv2} & - & TPAMI'25 & 78.6 & 84.8 & 80.7 & 89.2 & 83.3 & 39.4 & 36.0 \\
     & CoD$^2$~\cite{xiong2026cod2} & - & AAAI'26 & 80.1 & 86.9 & 81.6 & 90.9 & 84.9 & -- & -- \\
     & FoundationGait~\cite{ye2026silhouette} & - & ECCV'26 & 84.0 & 87.5 & 85.5 & 91.7 & 87.2 & 53.3 & 48.0 \\
    \midrule
    \multirow{2}{*}{Sil.+Ske.} & BiFusion~\cite{peng2024bifusion} & - & MTAP'24 & 62.6 & 67.6 & 66.3 & 66.0 & 65.6 & -- & -- \\
     & SkeletonGait++~\cite{fan2024skeletongait} & - & AAAI'24 & 79.1 & 83.9 & 81.7 & 89.9 & 83.7 & -- & -- \\
    Sil.+Par. & XGait~\cite{zheng2024xgait} & - & MM'24 & 72.8 & 77.0 & 79.1 & 80.5 & 77.4 & 41.5 & 38.1 \\
    Sil.+Par.+Flow & MultiGait++~\cite{jin2025multigait} & - & AAAI'25 & 83.9 & 89.0 & 86.0 & 91.5 & 87.6 & -- & -- \\
    \multirow{2}{*}{RGB+Sil.} & GaitEdge~\cite{liang2022gaitedge} & - & ECCV'22 & 66.9 & 74.0 & 70.6 & 77.1 & 72.2 & -- & -- \\
     & DenoisingGait~\cite{jin2025denoisinggait} & SD-1.5~\cite{rombach2022ldm} & CVPR'25 & 84.0 & 88.0 & 90.1 & 95.9 & 89.5 & -- & -- \\
    \midrule
    \multirow{8}{*}{RGB} & AP3D~\cite{gu2020ap3d} & - & ECCV'20 & 53.4 & 57.3 & 69.7 & 91.4 & 68.0 & -- & -- \\
     & PSTA~\cite{wang2021psta} & - & ICCV'21 & 42.2 & 52.2 & 60.3 & 84.5 & 59.8 & -- & -- \\
     & PiT~\cite{zang2022pit} & - & TII'22 & 41.0 & 47.6 & 64.3 & 91.0 & 61.0 & -- & -- \\
     & BigGait~\cite{ye2024biggait} & DINOv2-S~\cite{oquab2024dinov2} & CVPR'24 & 82.6 & 85.9 & 87.1 & 93.1 & 87.2 & 88.0 & -- \\
     & Gait-X~\cite{wang2025gaitx} & - & ICCV'25 & 82.0 & 86.0 & 90.7 & 95.7 & 88.6 & -- & -- \\
     & BiggerGait~\cite{ye2025biggergait} & DINOv2-S~\cite{oquab2024dinov2} & NeurIPS'25 & 89.8 & 92.1 & 93.7 & 97.5 & 93.3 & 85.7 & -- \\
     & LMGait~\cite{wu2026lmgait} & DINOv2-S~\cite{oquab2024dinov2} & AAAI'26 & 84.8 & 87.0 & 88.5 & 93.6 & 88.5 & -- & -- \\
     & GaitMax~\cite{huang2026gaitmax} & DINOv3-S~\cite{simeoni2025dinov3} & CVPR'26 & 86.6 & 88.2 & 90.2 & 93.2 & 89.6 & 83.6 & 74.2 \\
    \cmidrule{2-11}
     & \method & DINOv2-S~\cite{oquab2024dinov2} & Ours & \textbf{92.3} & \textbf{94.9} & \textbf{94.0} & \textbf{97.9} & \textbf{94.8} & \textbf{89.2} & \textbf{85.5} \\
    \bottomrule
  \end{tabular}}
\end{table*}

\subsection{Experimental setup}
\label{sec:exp_setup}

\paragraph{Datasets.}
We use six benchmarks that range from controlled gait datasets to ReID in the wild.
For gait, CCPG~\cite{li2023ccpg} has 200 identities with diverse clothing changes; CCGR-MINI, a subset of CCGR~\cite{zou2024ccgr}, has 970 identities under 53 covariates and 33 views, including steep pitch angles; CASIA-B*~\cite{yu2006casiab,liang2022gaitedge} has 124 identities in 11 views under normal, bag, and coat conditions; and SUSTech1K~\cite{shen2023lidargait} has 1,050 identities under clothing, carrying, occlusion, and night conditions.
For video-based ReID, CCVID~\cite{gu2022cal} has 226 identities in 2,856 cloth-changing sequences, and MEVID~\cite{davila2023mevid} has 158 identities captured by 33 cameras in the wild.
We train and test within CCPG, CCGR-MINI, and MEVID, and measure cross-domain generalization by training on CCGR-MINI and testing on CCPG, CASIA-B*, SUSTech1K, and CCVID without fine-tuning.

\paragraph{Protocols and metrics.}
We follow the official protocol and metrics of each benchmark, Rank-1 and mAP, and compare only with results reported under the same setting; on CCPG, same-view probe-gallery pairs are excluded~\cite{ye2025biggergait}.

\paragraph{Implementation details.}
For explicit imagination, Qwen-Image-2.1~\cite{wu2025qwenimage} renders the same outfit, side view, and back view of 8 frames per training sequence in 4 denoising steps.
Silhouettes come from official releases or BiRefNet~\cite{zheng2024birefnet}, and head close-ups are cropped around YOLO11-pose~\cite{jocher2024yolo11} keypoints.
On these renders, we train FoundationGait~\cite{ye2026silhouette} as the silhouette expert and BiggerGait~\cite{ye2025biggergait} as the other four experts.
For latent imagination, $G$ learns their targets for 10k iterations with AdamW, a learning rate of $2\times10^{-4}$, and a batch of 32 sequences of 4 frames.
For adaptive fusion, the condition heads are linear classifiers trained on view labels of CCGR-MINI and outfit labels of CCPG and MEVID.
Driven by their predictions, the condition table is fit by coordinate ascent, which tunes one entry at a time to maximize retrieval accuracy while penalizing its distance to the predefined table.
The condition table is fit on CCGR-MINI and CCPG, and refit on MEVID for within-domain ReID.
All networks are trained on two NVIDIA B300 GPUs.

\subsection{Comparison with state-of-the-art methods}
\label{sec:exp_sota}

\begin{table}[t]
  \centering
  \caption{Cross-domain gait recognition (Rank-1, \%). All recognizers are trained on the CCGR-MINI training set and directly tested on CCPG, CASIA-B*, and SUSTech1K.}
  \label{tab:sota_cross}
  \small
  \setlength{\tabcolsep}{2.5pt}
  \fittable{%
  \begin{tabular}{l cccc ccc cc}
    \toprule
    \multirow{2}{*}{Method} & \multicolumn{4}{c}{CCPG~\cite{li2023ccpg}} & \multicolumn{3}{c}{CASIA-B*~\cite{yu2006casiab}} & \multicolumn{2}{c}{SUSTech1K~\cite{shen2023lidargait}} \\
    \cmidrule(lr){2-5} \cmidrule(lr){6-8} \cmidrule(lr){9-10}
    & CL & UP & DN & BG & NM & BG & CL & CL & Overall \\
    \midrule
    GaitBase~\cite{fan2023opengait} & 20.8 & 31.3 & 38.3 & 70.2 & 66.3 & 57.7 & 33.9 & 28.7 & 48.0 \\
    DeepGaitV2~\cite{fan2023deepgaitv2} & 23.0 & 37.1 & 36.5 & 69.9 & 72.0 & 62.0 & 36.6 & 32.1 & 51.7 \\
    BigGait~\cite{ye2024biggait} & 22.9 & 42.2 & 25.0 & 80.5 & 90.1 & 87.9 & 58.4 & 71.9 & 85.6 \\
    BiggerGait~\cite{ye2025biggergait} & 23.2 & 44.5 & 29.5 & 86.7 & 97.2 & 96.4 & 66.7 & \textbf{82.1} & 93.8 \\
    \midrule
    \method & \textbf{39.4} & \textbf{65.2} & \textbf{41.2} & \textbf{95.8} & \textbf{98.6} & \textbf{98.1} & \textbf{79.2} & 80.5 & \textbf{95.0} \\
    \bottomrule
  \end{tabular}}
\end{table}

\begin{table}[t]
  \centering
  \caption{Cross-domain video-based ReID (\%). All models are tested directly on CCVID~\cite{gu2022cal}. CC: cloth-changing. LUP4M: LUPerson-4M~\cite{fu2021luperson}; Market-1501~\cite{zheng2015market}; LTCC~\cite{qian2020ltcc}; PRCC~\cite{yang2021prcc}.}
  \label{tab:sota_ccvid}
  \small
  \setlength{\tabcolsep}{1.9pt}
  \renewcommand{\arraystretch}{0.92}
  \fittable{%
  \begin{tabular}{lcc cc cc}
    \toprule
    \multirow{2}{*}{Method} & \multirow{2}{*}{Venue} & \multirow{2}{*}{Train data} & \multicolumn{2}{c}{General} & \multicolumn{2}{c}{CC} \\
    \cmidrule(lr){4-5} \cmidrule(lr){6-7}
    & & & R1 & mAP & R1 & mAP \\
    \midrule
    CAL~\cite{gu2022cal} & CVPR'22 & LTCC+PRCC & 74.8 & 29.4 & 73.9 & 26.7 \\
    SOLIDER~\cite{chen2023solider} & CVPR'23 & LUP4M+Market & 40.3 & 36.6 & 39.6 & 35.5 \\
    HAP~\cite{yuan2023hap} & NeurIPS'23 & LUP4M+Market & 54.7 & 45.1 & 52.4 & 41.3 \\
    CLIP3DReID~\cite{liu2024clip3dreid} & CVPR'24 & LTCC & 77.3 & 30.0 & 76.3 & 26.7 \\
    SapiensID~\cite{kim2025sapiensid} & CVPR'25 & WebBody4M & \textbf{92.6} & 77.8 & \textbf{88.7} & 72.2 \\
    BiggerGait~\cite{ye2025biggergait} & NeurIPS'25 & CCGR-MINI & 85.1 & 83.4 & 84.5 & 82.6 \\
    \midrule
    \method & Ours & CCGR-MINI & 87.4 & \textbf{86.1} & 86.9 & \textbf{85.5} \\
    \bottomrule
  \end{tabular}}
\end{table}

\begin{table}[t]
  \centering
  \caption{Within-domain video-based ReID (overall, \%). All models are trained and tested on MEVID~\cite{davila2023mevid}. Backbone: the pretrained encoder, or Ens. for the score fusion of AdaFace~\cite{kim2022adaface}, CAL~\cite{gu2022cal}, and AGRL~\cite{wu2020agrl}. Methods with far larger backbones~\cite{su2025hamobe,pathak2025csci} are not included in this table.}
  \label{tab:sota_mevid}
  \small
  \setlength{\tabcolsep}{3.8pt}
  \renewcommand{\arraystretch}{0.92}
  \coltable{%
  \begin{tabular}{lcc cc}
    \toprule
    Method & Venue & Backbone & mAP & R1 \\
    \midrule
    AGRL~\cite{wu2020agrl} & TIP'20 & ResNet-50~\cite{he2016resnet} & 19.1 & 48.4 \\
    AP3D~\cite{gu2020ap3d} & ECCV'20 & ResNet-50~\cite{he2016resnet} & 15.9 & 39.6 \\
    TCLNet~\cite{hou2020tclnet} & ECCV'20 & ResNet-50~\cite{he2016resnet} & 23.0 & 48.1 \\
    Attn-CL~\cite{pathak2020attncl} & AAAI'20 & ResNet-50~\cite{he2016resnet} & 18.6 & 42.1 \\
    BiCnet-TKS~\cite{hou2021bicnettks} & CVPR'21 & ResNet-50~\cite{he2016resnet} & 6.3 & 19.0 \\
    STMN~\cite{eom2021stmn} & ICCV'21 & ResNet-50~\cite{he2016resnet} & 11.3 & 31.0 \\
    PSTA~\cite{wang2021psta} & ICCV'21 & ResNet-50~\cite{he2016resnet} & 21.2 & 46.2 \\
    PiT~\cite{zang2022pit} & TII'22 & ViT-B/16~\cite{dosovitskiy2021vit} & 13.6 & 34.2 \\
    CAL~\cite{gu2022cal} & CVPR'22 & ResNet-50~\cite{he2016resnet} & 27.1 & 52.5 \\
    ShARc~\cite{zhu2024sharc} & WACV'24 & ResNet-50~\cite{he2016resnet} & 29.6 & 59.5 \\
    VCLIP~\cite{hill2025vclip} & WACV'25 & CLIP-B/16~\cite{radford2021clip} & 27.7 & 60.8 \\
    QME~\cite{zhu2025qme} & ICCV'25 & Ens. & 28.2 & 55.7 \\
    FusionAgent~\cite{zhu2026fusionagent} & CVPR'26 & Ens. & 28.7 & 54.7 \\
    \midrule
    \method & Ours & DINOv2-S~\cite{oquab2024dinov2} & \textbf{41.7} & \textbf{64.6} \\
    \bottomrule
  \end{tabular}}
\end{table}

\paragraph{Within-domain gait recognition.}
\method sets a new state of the art on both CCPG and CCGR-MINI (\cref{tab:sota_gait}).
Each family of prior methods has its own weakness.
Video-based ReID methods~\cite{gu2020ap3d,wang2021psta,zang2022pit} rely on appearance, so they cope with bag changes (BG) but break down under outfit changes (CL).
Silhouette-based methods discard body details, so even the gait foundation model FoundationGait~\cite{ye2026silhouette} reaches only 87.2\% mean accuracy on CCPG.
The strongest prior methods build on large vision models~\cite{ye2024biggait,ye2025biggergait,huang2026gaitmax}, yet even BiggerGait drops from 97.5\% under BG to 89.8\% under CL.
In contrast, \method raises the CL accuracy on CCPG by 2.5\% and the mean by 1.5\% over the best prior method, and reaches 89.2\% Rank-1 and 85.5\% mAP on CCGR-MINI, whose steep pitch angles change the viewpoint drastically.
By imagining people under canonical conditions, the identity world model handles both clothing and viewpoint changes, verifying it as a promising and robust paradigm for whole-body recognition.

\paragraph{Cross-domain gait recognition.}
The advantage of \method grows on unseen domains (\cref{tab:sota_cross}).
Prior methods learn invariance from the source data and degrade sharply under new clothing styles, scenes, and cameras; BiggerGait, for example, reaches only 23.2\% CL accuracy on CCPG.
\method lifts it to 39.4\% and outperforms the best prior method by 14.4\% and 5.2\% in mean accuracy on CCPG and CASIA-B* and by 1.2\% in overall accuracy on SUSTech1K.
Even on unseen domains, the identity world model still imagines useful canonical conditions and recognizes people reliably, verifying its strong generalization.

\paragraph{Cross-domain video-based ReID.}
Without fine-tuning, \method also achieves the best mAP for video-based ReID on CCVID (\cref{tab:sota_ccvid}).
Image-based cloth-changing methods~\cite{gu2022cal,liu2024clip3dreid} reach only about 30\% mAP on these videos, human-centric pretraining~\cite{chen2023solider,yuan2023hap} lifts it to at most 45.1\%, and SapiensID~\cite{kim2025sapiensid}, despite four million training images, has the highest Rank-1 but only 77.8\% mAP.
\method lifts its mAP by 2.7\% to 86.1\% (general) and by 2.9\% to 85.5\% (cloth-changing), as the imagined silhouette and same outfit carry body shape across outfits.
The identity world model thus excels at video-based ReID, verifying its versatility.

\paragraph{Within-domain video-based ReID.}
MEVID is more complex than walking gait benchmarks, as people in the wild change outfits, viewpoints, poses, and scales, often under occlusion.
Even so, \method outperforms all compared methods with small and base backbones (\cref{tab:sota_mevid}).
Video-based ReID methods~\cite{wu2020agrl,gu2020ap3d,hou2020tclnet,pathak2020attncl,hou2021bicnettks,eom2021stmn,wang2021psta,zang2022pit} and even the cloth-changing CAL~\cite{gu2022cal} rely on appearance and reach at most 27.1\% mAP, while shape-based~\cite{zhu2024sharc,liu2024clip3dreid}, vision-language~\cite{hill2025vclip}, and multi-expert~\cite{zhu2025qme,zhu2026fusionagent} methods improve robustness but still see each person only as captured.
\method improves the overall mAP by 12.1\% and \mbox{Rank-1} by 3.8\% over the best method in \cref{tab:sota_mevid}.
These results demonstrate that the identity world model extends beyond walking gait to unconstrained videos, reliably recognizing people who act freely in the wild.

\subsection{Quantitative and qualitative analysis}
\label{sec:exp_ablation}

\begin{table}[t]
  \centering
  \caption{Latent versus explicit imagination on CCPG (Rank-1, \%; 30 consecutive frames per sequence). Explicit: Qwen-Image-2.1~\cite{wu2025qwenimage} renders images in 4 steps, and YOLO11-pose~\cite{jocher2024yolo11} crops the head. FLOPs and latency are measured on a B300 in bf16.}
  \label{tab:latent_vs_explicit}
  \small
  \setlength{\tabcolsep}{3pt}
  \coltable{%
  \begin{tabular}{l cc cc}
    \toprule
    Method & CL & Mean & GFLOPs $\downarrow$ & Latency $\downarrow$ \\
    \midrule
    Without imagination & 81.1 & 86.6 & \textbf{91} & \textbf{1.2~\textit{ms}} \\
    Explicit imagination & \textbf{86.8} & \textbf{90.6} & 36{,}554 & 67.1~\textit{ms} \\
    \rowcolor{gray!15} Latent imagination (ours) & 86.0 & 90.1 & 528 & 5.9~\textit{ms} \\
    \bottomrule
  \end{tabular}}

  \vspace{8pt}
  \caption{Imagination target and training objective on CCPG (Rank-1, \%). (a)~Choice of imagination target. (b)~Regression versus flow matching. Defaults are shaded.}
  \label{tab:ablation}
  \setlength{\tabcolsep}{4.5pt}
  \coltable{%
  \begin{tabular}{l ccccc}
    \toprule
    Setting & CL & UP & DN & BG & Mean \\
    \midrule
    \multicolumn{6}{l}{\textit{(a) Imagination target}} \\
    \rowcolor{gray!15} \quad Early block (ours) & \textbf{92.3} & \textbf{94.9} & \textbf{94.0} & \textbf{97.9} & \textbf{94.8} \\
    \quad Late block & 90.9 & 94.2 & 93.1 & 97.8 & 94.0 \\
    \quad Final embedding & 85.5 & 89.5 & 90.0 & 97.0 & 90.5 \\
    \midrule
    \multicolumn{6}{l}{\textit{(b) Training objective}} \\
    \quad Flow matching & 91.8 & 94.3 & 93.5 & 97.8 & 94.3 \\
    \rowcolor{gray!15} \quad Regression (ours) & \textbf{92.3} & \textbf{94.9} & \textbf{94.0} & \textbf{97.9} & \textbf{94.8} \\
    \bottomrule
  \end{tabular}}
\end{table}

\paragraph{Latent versus explicit imagination.}
Latent imagination removes the main cost of imagination at test time (\cref{tab:latent_vs_explicit}).
Explicit imagination must render multiple images per frame and encode each again, costing 36.6 TFLOPs and 67.1\,ms per frame even with a 4-step distilled editor, cached prompts, and CUDA graphs.
\method imagines all conditions in a single forward pass with 528 GFLOPs at 5.9\,ms per frame, 69$\times$ fewer FLOPs and 11.4$\times$ faster.
Latent imagination thereby resolves the high latency challenge.

Beyond speed, latent imagination keeps most of the gain of explicit imagination.
On the same frames, rendering raises the mean accuracy from 86.6\% to 90.6\%, and latent imagination reaches 90.1\%, recovering 88\% of the gain at one eleventh of the latency.
Latent imagination is thus nearly as informative as rendering.

\begin{figure*}[t]
  \centering
  \includegraphics[width=\linewidth]{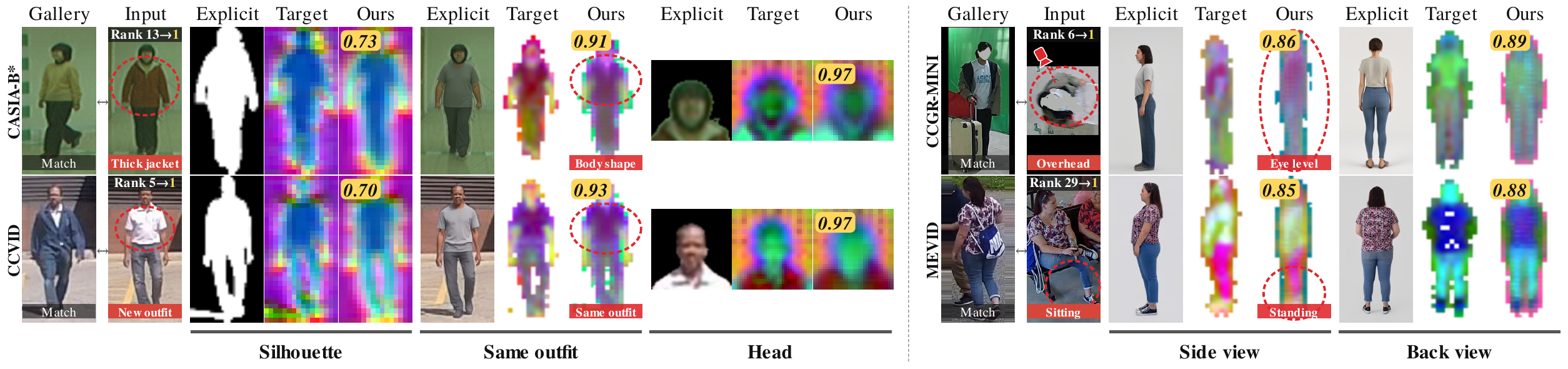}
  \definecolor{simyellow}{RGB}{255,215,94}
  \DeclareRobustCommand{\simbadge}[1]{\tikz[baseline=(b.base)]\node[fill=simyellow,rounded corners=1.5pt,inner xsep=1.5pt,inner ysep=0.8pt,text={rgb,255:red,17;green,17;blue,17}](b){\bfseries\itshape #1};}
  \vspace{-14pt}
  \caption{Hard cases from CASIA-B*~\cite{yu2006casiab}, CCVID~\cite{gu2022cal}, CCGR-MINI~\cite{zou2024ccgr}, and MEVID~\cite{davila2023mevid}. Gallery: true match. Explicit: rendered by Qwen-Image-2.1~\cite{wu2025qwenimage}. Target: expert features of Explicit. Ours: predicted by the imagination head. \simbadge{Numbers}: cosine similarity.}
  \label{fig:qualitative}
\end{figure*}

\paragraph{Choice of imagination target.}
Early feature maps are the best imagination targets (\cref{tab:ablation}a).
Imagining the final embedding of each expert reaches only 90.5\% mean accuracy, even below the 93.3\% without imagination, because such an abstract target is hard to predict from RGB.
An early block stays spatial and close to the pixels yet already encodes identity, so it is both predictable and discriminative, reaching 94.8\%, while a late block drops to 94.0\%.
As all variants share the same imagination head, the gain comes from what is imagined rather than extra capacity.

\paragraph{Regression versus flow matching.}
Direct regression is the better way to learn imagination (\cref{tab:ablation}b).
Flow matching~\cite{lipman2023flowmatching}, trained from Gaussian noise and sampled in 10 Euler steps, reaches 94.3\% mean accuracy, 0.5\% lower, as each target is nearly determined by the input and recognition datasets are too small to learn a full distribution.
Regression is more accurate and needs one step, not ten.

\begin{table}[t]
  \centering
  \caption{Utility of each imagined feature (Rank-1, \%). Baseline: recognition with the original feature alone. Imagined Condition: change when the baseline is fused with this imagined feature at a fixed weight. Overhead: the \mbox{90$^\circ$-pitch} probes of CCGR-MINI. The last three test sets are cross-domain.}
  \label{tab:conditions}
  \small
  \renewcommand{\arraystretch}{0.85}
  \setlength{\tabcolsep}{2.75pt}
  \newcommand{\gain}[1]{\textbf{+#1}}
  \newcommand{\drop}[1]{\textcolor{black!45}{$-$#1}}
  \coltable{%
  \begin{tabular}{l c cc cc c}
    \toprule
    \multirow{2}{*}{Test set} & \multirow{2}{*}{Baseline} & \multicolumn{5}{c}{Imagined Condition} \\
    \cmidrule(lr){3-7}
    & & Sil. & Outfit & Side & Back & Head \\
    \midrule
    CCPG & 93.3 & \gain{1.7} & \drop{0.2} & \drop{1.9} & \drop{2.0} & \drop{0.3} \\
    CCGR-MINI & 87.2 & \gain{0.2} & \drop{1.0} & \drop{1.6} & \drop{1.7} & \drop{3.0} \\
    \quad Overhead & 40.1 & \drop{21.9} & \drop{10.9} & \gain{5.7} & \gain{3.6} & \drop{2.6} \\
    MEVID & 54.1 & \textcolor{black!45}{+0.0} & \textcolor{black!45}{+0.0} & \gain{7.0} & \gain{6.3} & \gain{3.8} \\
    CASIA-B* & 86.4 & \gain{5.6} & \gain{1.2} & \drop{1.1} & \drop{0.8} & \drop{1.5} \\
    SUSTech1K & 93.7 & \gain{0.9} & \gain{0.1} & \drop{0.1} & \drop{0.1} & \drop{2.9} \\
    CCVID & 85.3 & \gain{3.1} & \gain{1.8} & \drop{1.4} & \drop{1.2} & \gain{1.3} \\
    \bottomrule
  \end{tabular}}
\end{table}

\begin{table}[t]
  \centering
  \caption{Adaptive versus fixed fusion (Rank-1, \%). Baseline fusions treat all probe-gallery pairs equally; ours adapts to each pair. Cross-domain: mean of the four test sets, CCPG~\cite{li2023ccpg}, \mbox{CASIA-B*}~\cite{yu2006casiab,liang2022gaitedge}, SUSTech1K~\cite{shen2023lidargait}, and CCVID~\cite{gu2022cal}.}
  \label{tab:selection}
  \small
  \renewcommand{\arraystretch}{0.85}
  \setlength{\tabcolsep}{2.6pt}
  \coltable{%
  \begin{tabular}{l ccc c}
    \toprule
    \multirow{2}{*}{Strategy} & \multicolumn{3}{c}{Within-domain} & \multirow{2}{*}{\shortstack{Cross-\\domain}} \\
    \cmidrule(lr){2-4}
    & CCPG & {\footnotesize CCGR-MINI} & MEVID & \\
    \midrule
    w/o imagination & 93.3 & 87.2 & 54.1 & 77.9 \\
    \midrule
    \multicolumn{5}{l}{\textit{Baseline fusions}} \\
    \quad Z-score~\cite{jain2005score} & 91.7 & 82.4 & 61.7 & 81.1 \\
    \quad Min-max~\cite{jain2005score} & 91.0 & 82.3 & 62.0 & 80.9 \\
    \quad Median-MAD~\cite{jain2005score} & 91.4 & 82.2 & 62.0 & 80.9 \\
    \quad RRF~\cite{cormack2009rrf} & 91.1 & 78.1 & 59.5 & 81.4 \\
    \midrule
    \rowcolor{gray!15} Adaptive (ours) & \textbf{94.8} & \textbf{89.2} & \textbf{64.6} & \textbf{83.7} \\
    \bottomrule
  \end{tabular}}
\end{table}

\paragraph{Utility of each imagined feature.}
Each imagined feature helps under a different kind of change (\cref{tab:conditions}).
The silhouette and same outfit help under outfit changes: the silhouette improves five of the six test sets, by up to 5.6\% on CASIA-B*, and the same outfit adds up to 1.8\% on cross-domain tests.
The side and back views help under large viewpoint changes, by up to 7.0\% on MEVID and 5.7\% on the overhead probes of CCGR-MINI, where the silhouette and same outfit fail.
The head close-up helps on MEVID and CCVID, where the head is large enough to see.
Yet each also hurts elsewhere, such as the views on eye-level datasets and the head on gait datasets.
No single imagined feature helps everywhere, and their utility varies even within CCGR-MINI, so the fusion weights must adapt to every probe-gallery pair.

\paragraph{Adaptive versus fixed fusion.}
Adaptive fusion turns these complementary strengths into consistent gains (\cref{tab:selection}).
Baseline fusions weight all pairs alike, so weak imagined features dilute the original feature, and all four fall below recognition without imagination on CCPG and CCGR-MINI, by up to 9.1\% on CCGR-MINI.
Adaptive fusion instead weights each condition by how the two observations differ, relying on the side and back views mainly across viewpoint gaps and raising the clothing-independent conditions under outfit changes.
As a result, it is best in every column, improving over the best baseline fusion by 3.1\% on CCPG, 6.8\% on CCGR-MINI, 2.6\% on MEVID, and 2.3\% across domains. Adaptive fusion thus resolves the second challenge of varying utility.

\paragraph{Feature map visualization.}
On hard cases from four test sets, the imagined feature maps, shown by PCA, closely match their targets (\cref{fig:qualitative}).
Under clothing changes, a thick jacket or a new outfit hides the body shape, yet the same outfit and head close-up reach cosine similarities of 0.91 to 0.97.
From an overhead camera or a sitting pose, the side and back views are imagined standing at eye level, at 0.85 to 0.89.
The silhouette is the hardest, at 0.70 and 0.73, yet keeps the body outline.
With these imagined features, the true match of every case rises to rank 1, from 13th, 5th, 6th, and 29th with the original feature alone.

\section{Conclusion}
\label{sec:concl}
We present \method, the first identity world model for whole-body human recognition, which anticipates variation instead of suppressing it.
Supervised by explicit imagination, it predicts in a single forward pass the early feature maps of five condition experts from RGB alone, and adaptively fuses these imagined identity features for every probe-gallery pair.
\method runs over 11$\times$ faster than explicit imagination while keeping most of its gain, and outperforms prior methods in nearly all settings of six gait and video-based ReID benchmarks.
Our analysis further shows that early feature maps are better imagination targets than final embeddings, and that direct regression learns them better than flow matching.
We hope that identity world models open a new path toward robust recognition in the wild.

\noindent\textbf{Limitations.}
(1)~Training still relies on time-consuming offline rendering for supervision.
(2)~Imagination covers five canonical conditions, and other conditions, such as a frontal face, are left for future work.
(3)~As the condition experts are sequence-based recognizers, evaluation mainly covers gait and video-based ReID benchmarks.

\noindent\textbf{Ethics.} All are publicly released datasets and models.

\FloatBarrier

\bibliographystyle{abbrvnat}
\bibliography{paper}

\end{document}